\documentclass{article}

\usepackage{microtype}
\usepackage{graphicx}
\usepackage{subfigure}
\usepackage{booktabs} 
\usepackage{diagbox}
\usepackage{hyperref}

\usepackage[accepted]{icml2025}

\usepackage{amsmath}
\usepackage{amssymb}
\usepackage{mathtools}
\usepackage{amsthm}

\usepackage[capitalize,noabbrev]{cleveref}

\theoremstyle{plain}

\theoremstyle{definition}

\theoremstyle{remark}

\usepackage[textsize=tiny]{todonotes}

\icmltitlerunning{COREVQA}

\begin{document}

\twocolumn[
\icmltitle{COREVQA: A Crowd Observation and Reasoning Entailment Visual Question
Answering Benchmark}

\begin{icmlauthorlist}
\icmlauthor{Ishant Chintapatla\textsuperscript{†*}}{}
\icmlauthor{Kazuma Choji\textsuperscript{†*}}{}
\icmlauthor{Naaisha Agarwal\textsuperscript{†*}}{}
\icmlauthor{Andrew Lin\textsuperscript{*}}{}
\icmlauthor{Hannah You\textsuperscript{*}}{}
\icmlauthor{Charles Duong\textsuperscript{°*}}{}
\icmlauthor{Kevin Zhu\textsuperscript{°*}}{}
\icmlauthor{Sean O'Brien\textsuperscript{°*}}{}
\icmlauthor{Vasu Sharma\textsuperscript{°*}}{}
\end{icmlauthorlist}

\begin{center}
\textsuperscript{†}Equal Contribution \quad
\textsuperscript{*}Algoverse AI Research \quad
\textsuperscript{°}Equal Senior Authorship \quad
\end{center}

\icmlcorrespondingauthor{Ishant Chintapatla}{ishantyunay@gmail.com}
\icmlcorrespondingauthor{Kazuma Choji}{kchoji@hmc.edu}\icmlcorrespondingauthor{Naaisha Agarwal}{naaishaagarwal@gmail.com}

\icmlkeywords{Visual Question Answering, Multimodal Bias, Machine Learning}

]

\printAffiliationsAndNotice{}

\begin{abstract}
Recently, many benchmarks and datasets have been developed to evaluate Vision-Language Models (VLMs) using visual question answering (VQA) pairs, and models have shown significant accuracy improvements. However, these benchmarks rarely test the model's ability to accurately complete visual entailment, for instance, accepting or refuting a hypothesis based on the image. To address this, we propose COREVQA (Crowd Observations and Reasoning Entailment), a benchmark of 5608 image and synthetically generated true/false statement pairs, with images derived from the CrowdHuman dataset \cite{shao2018crowdhuman}, to provoke visual entailment reasoning on challenging crowded images. Our results show that even the top-performing VLMs achieve accuracy below 80\%, with other models performing substantially worse (39.98\%-69.95\%). This significant performance gap reveals key limitations in VLMs’ ability to reason over certain types of image–question pairs in crowded scenes.

\footnote{Huggingface link: \href{https://huggingface.co/datasets/COREVQA2025/COREVQA}{\nolinkurl{https://huggingface.co/datasets/COREVQA2025/COREVQA}}}
\footnote{Github link: \href{https://github.com/corevqa/COREVQA}{\nolinkurl{https://github.com/corevqa/COREVQA}}}
\end{abstract}

\section{Introduction}
Vision language models (VLM), such as GPT-4.1 \cite{achiam2023gpt} and Gemini 2.5 Pro \cite{GoogleGemini2.5ProPreview0506}, have shown remarkable capabilities in image understanding and multimodal task completion \cite{li2025benchmark}. As VLMs grow more sophisticated, the demand for rigorous evaluation methods that assess deep visual and textual understanding becomes increasingly critical. \cite{huang2023language, agrawal2018don, goyal2017making}. 

However, despite offering a wide variety of tasks, existing VLM evaluation benchmarks often fall short in assessing nuanced understanding and deep comprehension of natural situations and analysis, primarily due to simple and easy-to-understand images/questions. These limitations mean models may succeed (and continue to succeed) by exploiting superficial cues or relying on parametric knowledge, without robust visual processing. The impact is impeding VLM improvements due to a scarcity of robust multimodal reasoning assessments \cite{li2025benchmark, guan2024hallusionbench}.

To fill this void in VLM assessment, we propose COREVQA (Crowd Observations and Reasoning Entailment Visual Question Answering)--a challenging evaluation benchmark based on images of dense human crowds in complex, natural settings. While existing crowd-based datasets only assess or focus on recognition, detection, and counting \cite{wang2020nwpu, wang2020panda, sindagi2020jhu, wang2019learning, zhang2016data, xie2019visual}, COREVQA requires models to integrate fine-grain visual analysis with textual logic in scenarios where visual ambiguity and seemingly trivial, easy-to-miss details are key. COREVQA aims to help researchers spot flaws and gaps in VLM understanding that will spur improvement in robustness.

Our main contributions are as follows.

\begin{itemize}
  \item We propose a pipeline to synthetically generate difficult questions for specific images based on typical VLM weaknesses.
  \item We created the first large-scale benchmark with multi-person, crowd-based images for evaluating VLM capability in busy scenarios (using our proposed pipeline).
  \item We evaluate several state-of-the-art (SOTA) VLMs on COREVQA, revealing across-the-board struggle with nuances and fine details when dealing with images with several people overflowing with diverse shapes, colors, and sizes.
\end{itemize}

\section{Related Work}

\subsection{Vision-Language Benchmarks}

Several benchmarks have become standard for evaluating core Visual-Question Answering (VQA) abilities.  VQAv2~\cite{goyal2017making} is the successor to the original VQA dataset \cite{antol2015vqa}, which aims to assess general VQA performance through a more balanced and challenging benchmark than VQA. Though they are still used for standardized evaluation, typical VQA datasets (like OK-VQA \cite{marino2019ok} and TextVQA \cite{singh2019towards}) lack complexity and difficulty. \cite{goyal2017making}. Newer datasets that analyze visual reasoning, understanding, recognition, and question answering include MMTBench \cite{ying2024mmt}, VCR \cite{zellers2019recognition}, MM-Vet \cite{yu2023mm}, SEEDBench \cite{li2023seed}, and NaturalBench \cite{li2024naturalbench}.  Most recent datasets overlook improving evaluation quality, and focus on assessing a wider range of tasks for easy and standardized comparison with other models \cite{fu2024mme}. These benchmarks include MMBench \cite{liu2024mmbench}, MMMU \cite{yue2024mmmu}, MMStar \cite{chen2024we}, M3GIA \cite{song2024m3gia}. Since most popular VLM benchmarks are designed in a multiple-choice format, further research is required to assess their robustness and reliability \cite{li2025benchmark}. Some others include NTSEBENCH \cite{pandya2024ntsebench} and VLDBENCH \cite{raza2025vldbench}, which target specific areas of visual reasoning. Difficult, targeted datasets like HallusionBench \cite{guan2024hallusionbench} have been created to evaluate key VLM weaknesses. 

Visual entailment benchmarks such as SNLI-VE \cite{xie2019visual}, Defeasible Visual Entailment \cite{zhang2025defeasible}, and VALSE \cite{parcalabescu2021valse} have all created questions that help test the models' ability to understand the text in regards to the image. 

Primary crowd-based datasets include NWPU-Crowd \cite{wang2020nwpu}, JHU-CROWD++ \cite{sindagi2020jhu}, PANDA \cite{wang2020panda}, and GCC \cite{wang2019learning}. 

By using a true/false format (rather than the primary usage of multiple choice \cite{li2025benchmark}) we reduce confounding variables found in answer choices because models cannot pick a correct choice when only given true or false (by nature) \cite{balepur2024artifacts}. Through our crowd-based dataset and synthetic generation we probe both visual entailment and textual comprehension, unlike many datasets mentioned that typically lack complex images (see more in \ref{sec:appendix-6}).

\section{COREVQA}
COREVQA is a novel VQA benchmark designed to evaluate VLMs' capabilities for detailed visual inspection and multi-step visual entailment. The benchmark features true/false statements about images that sound plausible but require careful visual grounding to verify their entailment status.

\subsection{Benchmark Overview}

COREVQA tests two core capabilities: depth of visual entailment and precision in analyzing fine-grained visual details. Models must classify the truthfulness of complex statements about images, with statements averaging 30 words and featuring complex sentence structures. The binary classification task assesses two main tasks -- Meticulous Visual Inspection: identifying subtle details in visual clutter or peripheral regions, and Complex Visual Entailment: understanding spatial relationships, making contextual inferences, and resisting plausible misdirection.

The benchmark has 5,608 unique image-statement pairs. Images come from the CrowdHuman dataset \cite{shao2018crowdhuman}, featuring diverse indoor and outdoor environments with groups of people. Each image is paired with one unique true/false statement generated through prompting ChatGPT for true statements and Claude for false statements. Ground truths were hand-labeled.

\begin{table}[H]
\centering
\caption{Key Statistics of the COREVQA Dataset}
\label{tab:dataset_stats}
\small
\begin{tabular}{ll}
\toprule
\textbf{Characteristic} & \textbf{Value} \\
\midrule
Dataset size & 5,608 image-statement pairs \\
true statements & 1,566 (27.9\%) \\
false statements & 4,042 (72.1\%) \\
Avg. statement length & 30.20 words \\
Statements w/ commas & 94.26\% \\
\bottomrule
\end{tabular}
\end{table}

\subsection{Data Collection}

\subsubsection{Image Sourcing}

Images were sourced from the CrowdHuman dataset \cite{shao2018crowdhuman} (\ref{sec:appendix-data-analysis}).

\subsubsection{True/False Statement Generation}

Statements were generated using ChatGPT 4.1 and Claude 3 Opus to increase diversity in the questions (see Appendix~\ref{sec:appendix-chose-model} for the reasoning behind this). Both were guided by an iteratively refined `true prompt' and `false prompt', designed to create tricky yet answerable statements. The prompt included three key components:
(1) a core directive that required generating statements that sounded natural while demanding close visual inspection; (2) content constraints that ensured each statement was grounded in visible elements; (3) a complexity requirement that each statement required chain-of-thought verification. (See Appendix~\ref{sec:appendix-prompts} for the full prompt.)

Statement generation strategies included leveraging subtle visual cues for true claims and introducing misleading or unsupported elements for false ones. A built-in self-reflection step prompted the model to justify how each statement might deceive, ensuring intentionality and grounded complexity (\ref{sec:appendix-4}). Our dataset maintains a slightly unbalanced distribution of true and false statements. (\ref{sec:appendix-Key Statistics of COREVQA Dataset}).

\subsubsection{Ensuring Question Quality}
When labeling the ground truths, we reprocessed any ambiguous questions or made minor grammar edits for clarity.

\subsection{Ground Truth}
Initially, we attempted to automate ground truth generation using a `solver prompt' applied to the same models that generated the statements. However, evaluation with a random sample of 250 image-statement pairs revealed only 89\% accuracy. To ensure complete accuracy of ground truths, the entire dataset was manually labeled. This ensures that all model scores (accuracy, precision, recall, and f1) are truly representative of their performance on COREVQA.

\subsection{Data Analysis}
\subsubsection{Images}
4,927 images (87.9\%) from the train01 split and 681 images (12.1\%) from the train02 split of the CrowdHuman dataset. These real-world photographs feature groups of people in diverse settings, providing a rich visual foundation for challenging visual entailment statements .

\subsubsection{Statements}

Statements exhibit significant syntactic complexity, frequent use of contrastive constructions (``while'': 32.9\%, ``despite'': 12.7\%). Content is people-centric, reflecting the CrowdHuman source, with common terms including ``person'' (47.3\% of statements), ``people'' (35.4\%), and actions like ``holding'' (46.7\%) and ``standing'' (19.5\%).

Over half (57.7\%) of statements use spatial terms, 39.0\% reference clothing, and 35.1\% mention color, highlighting the dataset's focus on detailed visual attributes and spatial understanding.

Notable biases include frequent mentions of specific objects: cameras (17.3\%), umbrellas (7.0\%), maps (1.1\%), and trophies (1.2\%). These patterns should be considered when evaluating model performance.

\subsection{Dataset Comparison}

Table~\ref{tab:dataset_comparison} compares COREVQA with other popular VLM benchmarks. While larger datasets like VQAv2 (1.1M samples) provide scale, our benchmark offers a more targeted evaluation through a focused combination of critical features absent in existing benchmarks. 

At 5.6K image-statement pairs, COREVQA balances comprehensive coverage with precision-targeted evaluation. This deliberate approach allows for more efficient and meaningful evaluation than would be possible with randomly sampled or user-generated content that test general capability.

COREVQA joins several other datasets in focusing on challenging questions on mulit-person imagery. These include NWPU-Crowd, which only evaluates counting and detection, HallusionBench \cite{guan2024hallusionbench}, which only focuses on adversarial examples, and SNLI-VE \cite{xie2019visual}, which utilizes simple imagery (\ref{sec:appendix-comparison}). COREVQA goes beyond these by providing a dataset with dense visual information and complex visual entailment that requires models to perform multi-step verification.

By strategically combining these dimensions, COREVQA offers unique diagnostic value in assessing VLMs' ability to perform the kind of careful visual verification required in high-stakes real-world applications. Our results reveal that even SOTA models with impressive performance on larger benchmarks struggle significantly when faced with this targeted combination of challenges.

\begin{table}[H]
\centering
\caption{COREVQA compared to existing benchmarks}
\label{tab:dataset_comparison}
\scriptsize
\begin{tabular}{lcccc}
\toprule
\textbf{Dataset} & \textbf{Size} & \textbf{Crowd Focus} & \textbf{Adversarial} & \textbf{Fine-grained} \\
\midrule
\textbf{COREVQA} & 5.6K & \textbf{Yes} & \textbf{Yes} & \textbf{Yes} \\
VQAv2 & \textbf{1.1M} & No & No & No \\
SNLI-VE & 565K & No & No & Partial \\
NWPU-Crowd & 5K & \textbf{Yes} & No & No \\
HallusionBench & 2K & No & \textbf{Yes} & No \\
MMBench & 2.9K & No & No & \textbf{Yes} \\
SEEDBench & 19K & No & No & \textbf{Yes} \\
\bottomrule
\end{tabular}
\end{table}

\section{Results and Analysis}

\subsection{Experimental Setup}
We evaluated GPT-4.1 \cite{openai2025gpt41}, GPT-4o mini \cite{achiam2023gpt}, Deepseek-Janus-Pro \cite{chen2025janus}, LLaVa-NeXT \cite{liu2024llavanext}, and Qwen2.5 vl 72b \cite{qwen2.5-VL} on COREVQA. All models tested received the same prompt to explicitly respond (to the image and T/F statement) with "True" or "False". We ran all models on the entire dataset, storing their responses. Given the binary classification nature of our task, accuracy is our main evaluation metric. We also report precision, recall, and F1 scores to provide a complete picture of model performance.

\subsection{Quantitative Results}
GPT4.1 achieves the highest overall accuracy (77.57\%) with GPT-4o Mini closely following, demonstrating a reasonable ability to verify both positive and negative claims. Janus Pro and Qwen2.5 vl 72b also do relatively well (72.31\% and 69.95\% respectively). However, Janus Pro has significantly low recall and F1 scores, indicating a strong bias toward answering false. LLaVa-NeXT displays near-perfect recall (99.68\%) but scores poorly on all other categories (\ref{tab:model_performance}). 

\begin{table}[H]
\caption{Model Performance on COREVQA}
\label{tab:model_performance}
\centering
\small
\resizebox{\columnwidth}{!}{%
\begin{tabular}{lcccc}
\toprule
\textbf{Model} & \textbf{Accuracy (\%)} & \textbf{Precision (\%)} & \textbf{Recall (\%)} & \textbf{F1 (\%)} \\
\midrule
GPT-4.1 & 77.57 & 57.36 & 76.63 & 65.60 \\
GPT-4o mini & 76.60 & 56.72 & 68.45 & 62.04 \\
Janus Pro & 72.31 & 64.44 & 1.85 & 3.60 \\
Qwen2.5 vl 72b & 69.95 & 47.91 & 87.23 & 61.85 \\
LLaVa-NeXT & 39.98 & 31.71 & 99.68 & 48.12 \\
\bottomrule
\end{tabular}
}
\end{table}

\subsection{Case Studies and Examples}

Figure 1 showcases a striking example where all tested models unanimously failed on what appears to be a straightforward visual assessment. Despite the seemingly precise description, all five models incorrectly classified this statement as true when the ground truth is false.

This example illustrates VLMs' tendency to generate plausible but factually incorrect interpretations when precise action recognition is required. The models likely detected people in the foreground and recognized basic poses, but failed to accurately discern if there was exactly one individual who was visibly holding a phone to their ear. See further examples (\ref{sec:appendix-sample-questions}). 

Broader dataset analysis revealed that 21.5\% of questions posed significant challenges for models, with at least two models providing incorrect answers. Within these difficult cases, we identified several consistent failure patterns such as not recognizing human actions, overlooking details, inaccurately counting, not understanding spatial relations, or not handling negation properly (\ref{sec:appendix-failure-patterns}).

COREVQA demonstrates that fine-grained verification in natural, complex images remains a significant challenge. These patterns reveal fundamental limitations in current VLMs' ability to perform the kind of meticulous visual verification required for real-world applications involving complex scenes. 
\begin{figure}[t]
\centering
\includegraphics[width=1.0\linewidth]{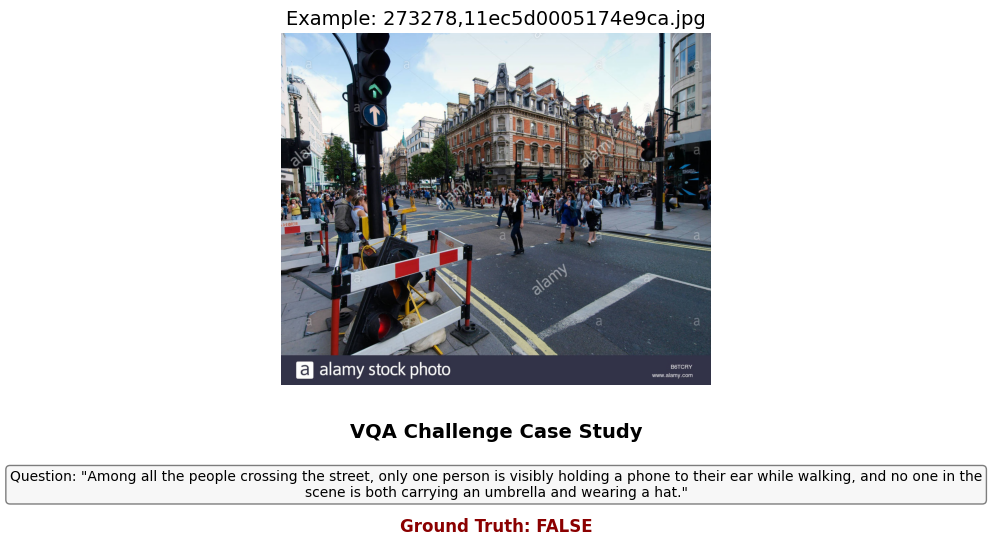}
\caption{Case study showcasing a challenging VQA scenario where all models (GPT-4.1, GPT-4o mini, JanusPro, LLaVA-NeXT, and Qwen) responded TRUE when the correct answer was FALSE. }
\label{fig:example_cases}
\end{figure}

\section{Limitations and Future Work}
\subsection{Current Limitations}
The requirement of human labeling prevents fast scaling. Furthermore, using a true or false format does not provide insight into key areas where a model went wrong (or right) when answering a question. Generating true and false questions solely with ChatGPT and Claude Opus also has the potential to introduce linguistic biases or limit the stylistic diversity of the statements. COREVQA has an uneven split of true and false questions, with there being more false than true.

\subsection{Suggested Directions for Improvement}
Currently, we evaluate only one high-parameter open-source model (Qwen) on COREVQA. We can test more models, such as InternVL3-78b, which are high-performing and open-source \cite{zhu2025internvl3}. COREVQA itself could be expanded upon by incorporating crowd data from various sources to increase image diversity. Further analysis and confidence metrics could be conducted to improve the reliability of model accuracy scores. One extension to this paper would be to fine-tune SOTA models on COREVQA and analyze possible improvements in general performance in popular benchmarks. Another extension would be to add non-human-centric images and questions, like groups of animals, to the dataset and analyze model performance.

\section{Conclusion}
This paper introduces COREVQA (Crowd Observations and Reasoning Entailment), a novel Visual Question Answering (VQA) benchmark designed to rigorously evaluate Vision-Language Models (VLMs). Existing VLM benchmarks often rely on simple images or questions, and existing crowd-based datasets exclusively focus on detection, recognition, and counting. Recognizing this, COREVQA was made with high-quality crowd images and synthetically generated tricky questions, targeting visual entailment capabilities where models must accurately verify or refute claims about image content. Our initial experiments with several state-of-the-art VLMs show that even the most capable models struggle with the complexity of our benchmark. Through COREVQA we aim to expose more gaps in VLM integration of visual understanding and logical analysis, paving the way for further improvements.

\bibliography{sources}
\bibliographystyle{icml2025}

\appendix
\section{Appendix}
\label{sec:appendix}

\vspace{0.5em}  

\subsection{Full Generation Prompts}
\label{sec:appendix-prompts}

\subsubsection*{False Prompt}
You are writing a False statement about a given image. The statement should sound completely natural, confident, and plausible — but require close visual inspection to determine if it’s actually correct.
Your goal is to trick a model that relies on language, assumptions, or common sense — and force it to use detailed visual and contextual reasoning.
Do **not** invent things that are fully out of frame, completely occluded, or dependent on unreadable text. Avoid using color as the main trick — models are strong at detecting color.
All questions must:
- Be complex and require chain-of-thought to answer. The complexity should come from the lengthy and advanced thought process required to answer the image, NOT the detail of the statement itself.
**Important Strategy Choices.**:
:x: When writing a FALSE Statement:
Create a statement that seems plausible but is visually incorrect. The goal is to bait the model into trusting assumptions, not what’s actually in the image.
Implied Physical Stability — Suggest ambiguous leaning/support just based off proximity, not grounded in visual details.
Quantifier Bait — Use vague or tricky counts (e.g. “at least three,” “only one”) to confuse object detection.
Intention/Action Fallacy — Imply an impossible or contradictory action (e.g. someone trying to do something that seems right, but in context shows is not feasible).
Similar Object Confusion — Identify an object that resembles and is used in context of a similar object but is incorrectly labeled.
Occlusion Trap — Imply something is visible and confirmable, when it's actually hidden or blocked.
Causal Mislead — Imply a cause-effect relationship that isn't actually supported by the image context (e.g. “because X is happening, Y must be true”).
Back-to-Front Error — Describe an event in the background as if it’s in the foreground, causing false assumptions about prominence or relevance.
Non-Functional Object Claim — Attribute an incorrect purpose to an object (e.g. assuming a pipe is a handrail, or a bag is a tool).
Time-Based Confusion — Describe a sequence-dependent state that cannot be confirmed from a single frame (e.g. “this person has just finished climbing”).
Schema Reversal: Use roles that *sound expected* (e.g. “a coach giving a trophy”) but flip them visually (e.g. it’s actually a parent or official).
Implied Grouping: Suggest affiliation between people (e.g. teammates, family) based on position or dress — even if incorrect.
Hidden Contradictions: Embed one small, one or two word, subtle error (e.g. missing ID badge, wrong uniform, incorrect object) inside an otherwise believable sentence.
Overloaded Compositions: Combine multiple assumptions (e.g. action + role + object) so the model must validate *each part* to avoid error.
---
Before responding, analyze your initial response. Make sure it is visually grounded. Also ask, how can I trick someone? How can I trick them into thinking the WRONG answer? 

\subsubsection*{True Prompt}
Generate a true statement about this image that requires sophisticated visual entailment to verify. Format your response as:
**STATEMENT:** [Your single true statement about the image]
**REASONING:** [Brief explanation of why this statement requires careful visual analysis]
The statement should:
1. Be concise and ABSOLUTELY factually verifiable based on visual evidence
2. Employ ONE of these reasoning approaches:
   - **Spatial reasoning**: Describe precise relationships between multiple elements, their arrangement patterns, or compositional structure
   - **Temporal/causal inference**: Identify evidence of what just happened or is about to happen based on visual cues of action, intent, or sequence
   - **Background knowledge integration**: Connect visible details with common knowledge about contexts, functions, or conventions to draw valid conclusions
3. Challenge verification by:
   - Requiring attention to multiple visual elements simultaneously
   - Pointing out non-obvious patterns or relationships
   - Making valid inferences that go beyond listing visible objects
4. Avoid statements that:
   - Are immediately obvious from a glance
   - Merely describe the presence of common objects
   - Could apply to many similar images
   - Require speculation beyond what's visually evident
   - Make universal claims about all people or objects in the image
IMPORTANT: Double check to make sure each claim in the statement is 100\% visually true without exception. When in doubt, use more conservative qualifiers.

\subsection{Key Statistics of the COREVQA Dataset}
\label{sec:appendix-Key Statistics of COREVQA Dataset}

\subsection{Prompt Strategies}
\label{sec:appendix-4}
For true statements, strategies included describing details buried in visual clutter, peripheral elements, subtle interactions, or surprising but correct conclusions. For false statements, strategies included implying unsupported physical states, using confusing quantifiers, mislabeling similar objects, suggesting invisible occluded elements, or creating back-to-front errors. The prompt included a self-reflection step for the LLM to analyze how the statement might deceive, ensuring intentional and grounded trickiness.

\subsection{Choosing Models For Question Generation}
\label{sec:appendix-chose-model}
When initially deciding which model we would use to create questions, we tested Gemini 2.5 Pro, Claude Opus, and GPT-4.1 on a generator prompt for both true and false questions. We ran through random samples of generations from each model and found that true statements from GPT-4.1 and false statements from Opus were most difficult for other models to answer. After this, we built up the generator prompt with patterns and details we saw in these generations.

\subsection{COREVQA compared to existing benchmarks}
\label{sec:appendix-comparison}
\paragraph{Crowd-Based Visual Complexity} COREVQA joins NWPU-Crowd in focusing on challenging multi-person imagery, but with a crucial difference: while NWPU-Crowd addresses only counting and detection, our benchmark evaluates complex visual entailment in these scenes. This requires models to parse dense visual information with multiple subjects, occlusions, and interactions—a fundamental capability for real-world applications.

\paragraph{Adversarial Design Philosophy} Like HallusionBench \cite{guan2024hallusionbench}, our benchmark employs adversarial examples, but we specifically target the intersection of visual complexity and linguistic verification. Each statement is systematically designed to sound plausible while requiring careful inspection to verify or refute. This approach specifically targets the critical gap between language understanding and visual grounding.

\paragraph{Fine-Grained Visual Entailment} Most distinctively, COREVQA is the only benchmark specializing in fine-grained visual entailment that demands precise visual grounding in complex scenes. While SNLI-VE \cite{xie2019visual} offers partial fine-grained verification, it uses primarily simpler imagery and lacks our systematic adversarial design. COREVQA requires models to perform multi-step verification by decomposing complex claims and meticulously verifying each component against visual evidence.

\subsection{Further Description of COREVQA}
\label{sec:appendix-6}
Rather than evaluating various diverse tasks, or exclusively focusing on one aspect of performance, like text recognition or hallucination, COREVQA combines visual entailment and textual comprehension with heavy occlusion from our crowd-based images \cite{xie2019visual}. This combination takes the difficult aspects from existing benchmarks and combines them with a focus on crowds to provide quality, in-depth assessment that can be generalizable towards real-world scenarios.

LLMs display an unusual ability to select correct (multiple choice) answers without access to their respective questions, which may reflect the exploitation of dataset artifacts or successful inference from choices alone, rather than a complete understanding of the question and choices together \cite{balepur2024artifacts}. The same capability applies to VLMs because they have an LLM component \cite{li2025benchmark}. By nature, true or false answer choices are insufficient to infer a question from or attach semantic context. Therefore, COREVQA improves upon current benchmarks by removing confounding variables in evaluation.  

Though they are high-quality and ideal for evaluation datasets, crowd images have yet to be utilized towards assessing VLM performance outside of recognition, detection, or counting. These datasets either lack questions to go along with their images, or the questions revolve around human detection or counting. Our synthetic data generation pipeline serves as a tool to bring forth new datasets when images are available, but creating questions would be costly or time-consuming. There is currently a lack in high-quality datasets for VLMs for which our pipeline contributes toward a solution. 

\subsection{Data Analysis}
\label{sec:appendix-data-analysis}

\subsubsection{Initial Model Assessment}

Our evaluation of five SOTA VLMs on COREVQA reveals significant performance gaps and distinctive response patterns that highlight the challenging nature of our benchmark.

\subsubsection{Failure Patterns}
\label{sec:appendix-failure-patterns}
\textbf{Action Recognition Failures:} The most prevalent pattern (appearing in 81.3\% of difficult cases) involved incorrectly identifying or interpreting human actions and behaviors. Models struggled to distinguish between similar actions or accurately recognize specific gestures, particularly in crowded scenes with multiple subjects.
\textbf{Detail Oversight:} In 78.1\% of challenging questions, models missed crucial visual details essential for accurate assessment. This was especially pronounced in lengthy, detailed descriptions (over 150 characters) that required verifying multiple visual elements simultaneously.
\textbf{Counting Inaccuracies:} Quantification problems appeared in 60.8\% of difficult cases. Models frequently miscounted people, objects, or attributes, particularly when faced with partial occlusions or when distinguishing between similar items was required.
\textbf{Spatial Reasoning Failures:} Complex spatial relationships (e.g., "left of," "between," "behind") were misinterpreted in 41.7\% of challenging cases. Models showed particular weakness in tracking relative positions among multiple subjects, especially when referencing individuals in the background or periphery.
\textbf{Negation Handling:} Questions involving negation (31.3\% of difficult cases) proved consistently problematic. Models struggled to verify statements about what was \textit{not} present or true in an image, suggesting a bias toward confirmation rather than falsification.

\textbf{Model Performance} 
Performance varied significantly across models on these challenging cases. Janus Pro demonstrated the highest accuracy (64.0\%), followed by LLaVA-NeXT (35.6\%), while GPT-4.1 and GPT-4o mini performed notably worse (0.3\% and 0.2\%). This substantial performance gap suggests fundamentally different approaches to visual reasoning among these architectures, with particular implications for handling complex compositional questions about human behavior and interactions.
\subsection{Sample of Tricky Questions}
\label{sec:appendix-sample-questions}
The following two questions are some of the most difficult in the entire dataset, where all models got the answer wrong.
\begin{figure}[htbp]
\centering
\includegraphics[width=0.8\linewidth]{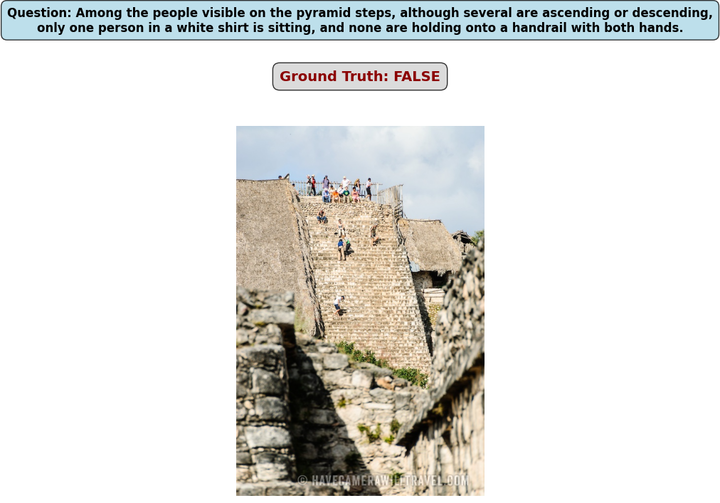}
\caption{Agreed (Incorrect) Model Answer: TRUE}
\label{fig:example_cases}
\end{figure}
\begin{figure}[htbp]
\centering
\includegraphics[width=0.8\linewidth]{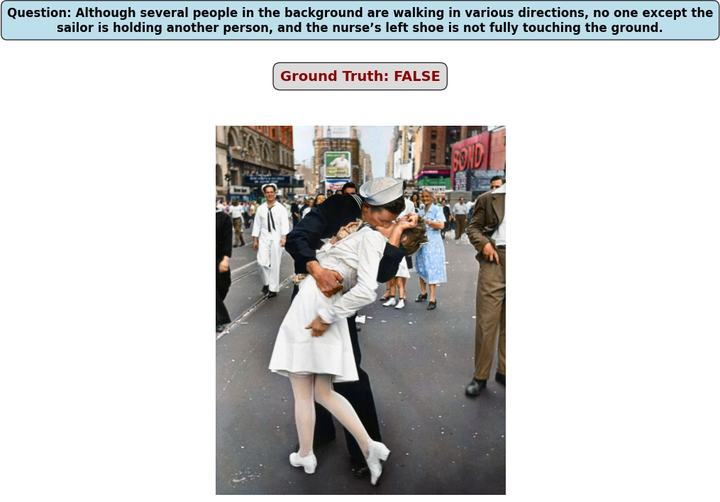}
\caption{Agreed (Incorrect) Model Answer: TRUE}
\label{fig:example_cases}
\end{figure}

\end{document}